\documentclass[runningheads]{llncs}

\usepackage{eccv}

\usepackage{eccvabbrv}
\usepackage{multirow}
\usepackage{graphicx}
\usepackage{booktabs}
\usepackage[accsupp]{axessibility}  

\usepackage{hyperref}

\usepackage{orcidlink}

\begin{document}
\title{Unmasking Bias in Diffusion Model Training}


\author{Hu Yu\inst{1} \and
Li Shen\inst{2} \and
Jie Huang\inst{1} \and
Hongsheng Li\inst{3} \and
Feng Zhao\inst{1} \thanks{Corresponding author.}}
\authorrunning{Hu Yu et al.}

\institute{MoE Key Laboratory of Brain-inspired Intelligent Perception and Cognition, University of Science and Technology of China \and
Alibaba Group \and The Chinese University of Hong Kong \\
\email{yuhu520@mail.ustc.edu.cn, fzhao956@ustc.edu.cn}
}

\maketitle

\begin{abstract}
Denoising diffusion models have emerged as a dominant approach for image generation, however they still suffer from slow convergence in training and color shift issues in sampling. 
In this paper, we identify that these obstacles can be largely attributed to bias and suboptimality inherent in the default training paradigm of diffusion models.
Specifically, we offer theoretical insights that the prevailing constant loss weight strategy in $\epsilon$-prediction of diffusion models leads to biased estimation during the training phase, hindering accurate estimations of original images. To address the issue, we propose a simple but effective weighting strategy derived from the unlocked biased part. Furthermore, we conduct a comprehensive and systematic exploration, unraveling the inherent bias problem in terms of its existence, impact and underlying reasons. These analyses contribute to advancing the understanding of diffusion models. Empirical results demonstrate that our method remarkably elevates sample quality and displays improved efficiency in both training and sampling processes, by only adjusting loss weighting strategy. The code is released publicly at \url{https://github.com/yuhuUSTC/Debias}

\end{abstract}

\section{Introduction}

\begin{figure}[t]
	\begin{center}
		\includegraphics[width=0.98\linewidth]{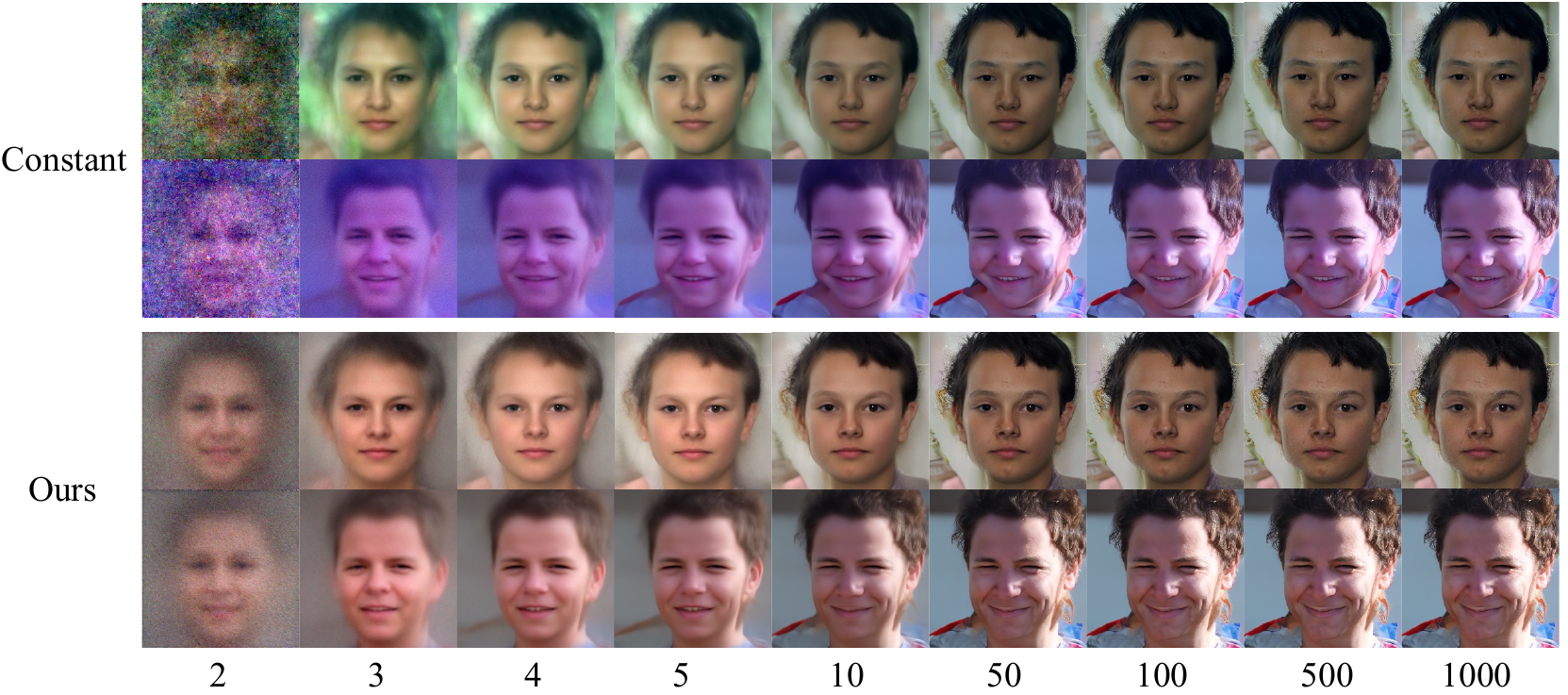}
	\end{center}
        \setlength{\abovecaptionskip}{-0.2cm}
        \setlength{\belowcaptionskip}{-0.6cm}
	\caption{Examples for the bias problem in $\epsilon$-prediction with constant weighting. Images are generated with different total sampling steps $T$. The upper two rows showcase samples obtained through constant weighting, exhibiting color shift and poor details. The bottom ones display samples generated using our method.}
	\label{fig:fewer_steps}
\end{figure}

Diffusion models~\cite{sohl2015deep,ho2020denoising} have emerged as powerful generative models that garner significant attention recently. Their popularity stems from the remarkable ability to generate diverse and high-quality samples~\cite{dhariwal2021diffusion,rombach2022high,ramesh2022hierarchical,nichol2021improved} as well as the training-stable loss form, compared to the adversarial training paradigms used in Generative Adversarial Networks (GANs) \cite{goodfellow2014generative}.  
Diffusion models often serve as a fundamental block and have exhibited impressive success on numerous tasks \cite{ruiz2023dreambooth, saharia2022image, yu2023high, yu2024uncovering}. While, it is usually employed as a black-box component in these works. 

There have been some attempts to delve into the methodology of diffusion models. The works in \cite{song2020denoising,liu2021pseudo,lu2022dpm,salimans2021progressive,meng2023distillation} target on the acceleration of the reverse sampling process. 
An alternative line of research has directed its attention towards the training objective, traditionally characterized by an elegantly simple loss function, i.e., the pixel-wise loss with a constant weight between the Gaussian noise and the predicted outcome as follows:
\begin{equation}
    \label{eq:1}
    L=\sum _t \mathbb{E}_{x_0,\epsilon}\left[||\epsilon-\epsilon_\theta(x_t, t)||^2\right].
\end{equation}
Prior works find that this loss formulation is less effective for training diffusion models, and alternative training objectives and weighting strategies are thus proposed. For instance, $\epsilon$-prediction with a range of customized weighting strategies \cite{choi2022perception,mardani2023variational,hang2023efficient}, or combining the strength of $\epsilon$-prediction and $x_0$-prediction to get new training targets 
\cite{salimans2021progressive, karras2022elucidating} can enhance model performance. However, a comprehensive examination of the underlying reasons and issues within the basic $\epsilon$-prediction in Eq. \ref{eq:1} is still lacking.

In this paper, we aim to fill this gap by conducting a detailed analysis to elucidate the bias and flaws associated with the basic $\epsilon$-prediction with constant weighting. Specifically, we provide a theoretical demonstration of its suboptimality, revealing its potential to introduce biased estimations during training and consequently diminish the overall performance of the model (as shown in Fig.~\ref{fig:fewer_steps}). To address the issue, we propose a simple but effective loss weighting strategy, termed the inverse of the Signal-to-Noise Ratio (SNR)'s square root, which is motivated from the uncovered biased part. Furthermore, we figure out several pivotal questions essential for systematically understanding the bias problem in conventional diffusion models, covering aspects of its existence, impact and underlying reasons.
Firstly, we demonstrate the existence of a biased estimation problem during the training process. The denoising network estimation may closely approach the target Gaussian noise at every step $t$, while, the corresponding estimated $\hat{x}_0$ may significantly deviate from the true $x_0$, with this deviation amplifying as $t$ increases.
Next, we analyse the influence of this biased estimation problem on the sampling process, termed as \textit{biased generation}. Biased generation primarily contributes to chaos and inconsistency in the early few sampling steps (left column of Fig.~\ref{fig:fewer_steps}), further affecting the final generation with error propagation effect.
We further uncover the root causes of biased estimation, elucidating that the importance and optimization difficulty of the denoising network vary significantly at different step $t$. 

We empirically show that the proposed method is capable of addressing the above problems and substantially elevates sample quality. The method can achieve superior performance to constant-weighting strategy with much less training iterations and sampling steps. Through comprehensive analyses and comparison, we also provide a unified prospective on existing weighting strategies ~\cite{choi2022perception,mardani2023variational,hang2023efficient}, highlighting the benefit of employing appropriate weights for loss penalties at different timesteps.

\section{Background and Related Work}\label{background}
\subsection{Preliminary of Diffusion models}
\noindent \textbf{Definition.}
Diffusion models~\cite{sohl2015deep,ho2020denoising} transform complex data distribution into simple noise distribution and learn to recover data from noise. 
The \textit{forward diffusion process} starts from a clean data sample $x_0$ and repeatedly injects Gaussian noise according to the transition kernel $q(x_t|x_{t-1})$ as follows:
\begin{equation}
	\label{eq:2}
	q(x_t|x_{t-1}) = N(x_t;\sqrt{1-\beta_t}x_{t-1},\beta_t I).  \\
\end{equation}
We can further derive closed-form expressions of distribution $q(x_t|x_0)$.
\begin{equation}
\label{eq:3}
    x_t=\sqrt{\alpha _{t}}x_{0} + \sqrt{1-\alpha _{t}}\epsilon,
\end{equation}
where $\epsilon\sim \mathcal{N}(0,\mathbf{I})$ and $\alpha_{t}:=\prod_{s=1}^t (1-\beta _{s})$.

The \textit{reverse denoise process} is trained to reverse the forward diffusion process in Eq.~\ref{eq:2} by learning the denoise network. Kingma et al.~\cite{kingma2021variational} further proposed the use of \textit{signal-to-noise ratio} (SNR) to simplify the representation the noise schedules in diffusion models, which is expressed as:
\begin{equation}
\label{eq:5}
    \text{SNR}(t)=\alpha _{t}/(1-\alpha _{t}).
\end{equation}

\noindent \textbf{Training objectives.}
Diffusion models are trained by optimizing a variational lower bound (VLB). For each step $t$, the denoising score matching loss $L_t$ is the distance between two Gaussian distributions, rewritten as:
\begin{equation}
\label{eq:6}
    \begin{aligned}
    L_{t}
    &=D_{KL}(q(x_{t-1}|x_t,x_0)~||~p_\theta(x_{t-1}|x_t)), \\
    &=\mathbb{E}_{x_0,\epsilon}\left[\frac{1}{2 \sigma_{t}^{2}}\left\|{C_1}x_0 +  {C_2} x_t  -\boldsymbol{x}_{\theta}\left(\mathbf{x}_{t}, 
    t\right)\right\|^{2}\right], \\
    &=\mathbb{E}_{x_0,\epsilon}\left[\frac{\beta_t^2}{(1-\beta _t)(1-\alpha _t)}||\epsilon-\epsilon_\theta(x_t, t)||^2\right]. \\
    \end{aligned}
\end{equation}
The expression of $C_1$ and $C_2$ as well as full derivation are available in the supplementary material. The denoising network is indeed optimized to approach $x_0$, while $\epsilon$ can also be employed as training target with a deterministic relationship to $x_0$. Ho et al. \cite{ho2020denoising} empirically demonstrated that $\epsilon$-prediction outperforms $x_0$-prediction. Additionally, they observed that the simplified objective (Eq.~\ref{eq:1}) with constant weight yields better sample quality, which subsequently becomes the default training objective of diffusion models.

\subsection{Related Work}
\noindent \textbf{Different training objectives.}
Many existing works adhere to the prevailing training objective in Eq.~\ref{eq:1}.
Recent methods \cite{choi2022perception,mardani2023variational,karras2022elucidating,mardani2023variational,salimans2021progressive,hang2023efficient} find Eq.~\ref{eq:1} less effective in performance and investigate improved training objectives and weighting strategies. They can be categorized into two types. One is $\epsilon$-prediction with various weighting strategies. Particularly, P2 \cite{choi2022perception} proposed a weighting strategy that prioritizes higher noise levels for recovering content information. 
Min-SNR \cite{hang2023efficient} interpreted the training goal from the perspective of multitask learning and studied the weighting strategy of Min-SNR. 
The other is combining the strengths of $\epsilon$-prediction and $x_0$-prediction to get new training targets \cite{salimans2021progressive, karras2022elucidating}. Salimans et al. \cite{salimans2021progressive} presented $v$-prediction for distilling diffusion models.
EDM \cite{karras2022elucidating} also realizes that directly predicting the Gaussian noise induces error amplification. While EDM resorts to precondition technique requiring network inputs and training targets to have unit variance, which is in the same spirit as previous reparameterization method like $v$-prediction to adaptively mix signal and noise.

\noindent \textbf{Induced color shift issues.}
Generated images of diffusion models suffer from errors in their spatial means, i.e., color shift. Song et al. \cite{song2020improved} observed this issue in the images generated by diffusion models, especially at higher image resolutions. 
They employ the exponential moving average strategy to alleviate this problem.
Deck et al. \cite{song2020improved} proposed a nonlinear bypass connection in the network to predict the mean of the score function.
P2 \cite{choi2022perception} suggests weighting the loss function to mitigate color shift, based on the intuition that crucial spatial features are generated early in the sampling process. Although the color shift issue can be alleviated by using these methods, they still face the challenges in interpreting the root cause of this phenomenon. In this paper, we unveil that the utilization of constant weights during the training stage plays a crucial role in causing the color shift problem, and this issue can be effectively addressed with our strategy.

\section{Theoretical Exploration of the Inherent Bias}   \label{section:theoretical}

\begin{figure*}[t]
	\begin{center}
		\includegraphics[width=0.9\linewidth]{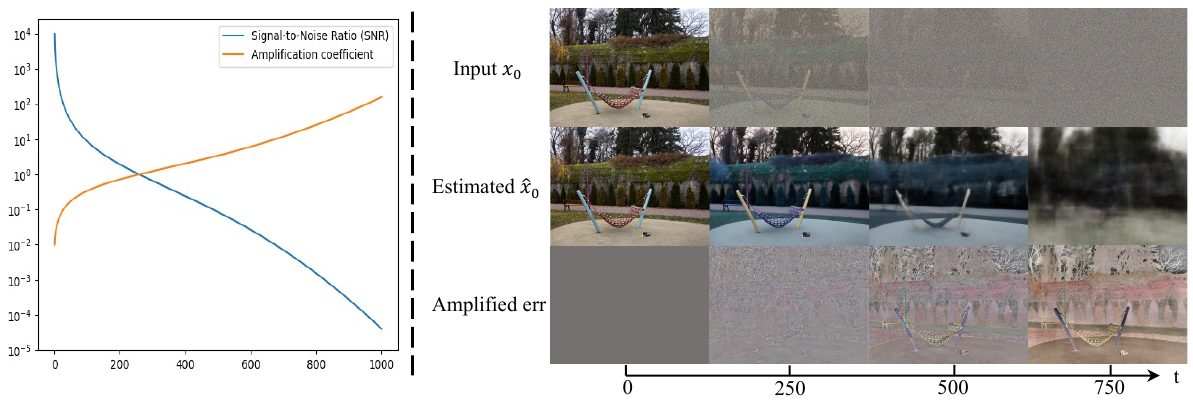}
	\end{center}
        \setlength{\abovecaptionskip}{-0.2cm}
        \setlength{\belowcaptionskip}{-0.4cm}
	\caption{Left: The visualization of SNR($t$) and amplification coefficient $\frac{1}{\sqrt{\text{SNR($t$)}}}$ at different timesteps. Right: The upper row is the input $x_t$ at different timesteps. We employ the diffusion model \cite{dhariwal2021diffusion} pretrained on ImageNet dataset to obtain the $\mathit{estimated \ \hat{x}_0}$ part and $\mathit{amplified \ error}$ part of each input $x_t$. The second row is the $\mathit{estimated \ \hat{x}_0}$. The bottom row is the corresponding $\mathit{amplified \ error}$ part. Apparently, as step $t$ gets larger, the $\mathit{estimated \ \hat{x}_0}$ severely deviates from $x_0$ and the $\mathit{amplified \ error}$ part gradually approaches $x_0$.}
	\label{fig:main}
\end{figure*}

\subsection{Constant Weighting Induces Bias in Training}
We treat $\epsilon$ as the explicit and direct target, and $x_0$ as the implicit but intrinsic target. 
Given the predicted noise $\epsilon_\theta(x_t, t)$ of the denoising network, we can easily derive the corresponding $\hat{x}_{0}$ from Eq.~\ref{eq:3} as follows:

\begin{equation}
    \label{eq:7}
    \begin{aligned}
    \hat{x}_{0} &= \frac{1}{\sqrt{\alpha _{t}}} x_{t} - \frac{\sqrt{1-\alpha _{t}}}{\sqrt{\alpha _{t}}}  \epsilon_\theta(x_t, t) \\
    &= \frac{1}{\sqrt{\alpha _{t}}} (\sqrt{\alpha _{t}}x_{0} + \sqrt{1-\alpha _{t}}\epsilon) - \frac{\sqrt{1-\alpha _{t}}}{\sqrt{\alpha _{t}}}  \epsilon_\theta(x_t, t) \\
    &= x_{0} + \frac{1}{\sqrt{SNR(t)}} (\epsilon - \epsilon_\theta(x_t, t)). \\
    \end{aligned}
\end{equation}

It is noticeable that while certain prior methods may reach the same derivation \cite{mardani2023variational}, they conclude the exploration at this point. Besides, instead of proceeding to train diffusion models for image generation, they apply it to other use cases. 
In stark contrast, this derivation is the start of our paper.
We conduct comprehensive analyses and studies to thoroughly unlock the problems behind this formulation, and propose a simple but effective solution.

Further, we can rewrite Eq.~\ref{eq:7} to express $x_0$ in terms of two components: the $\mathit{estimated \ \hat{x}_0}$ part and the $\mathit{amplified \ error}$ part.
\begin{equation}
\label{eq:8}
    x_0 = \underbrace{\hat{x}_{0}}_{estimated \ \hat{x}_0} + \underbrace{\frac{1}{\sqrt{SNR(t)}} (\epsilon_\theta(x_t, t) - \epsilon)}_{amplified \ error}.
\end{equation}

Although the difference between the predicted $\epsilon_\theta(x_t, t)$ and the target Gaussian noise $\epsilon$ may be very small at every step, the amplification coefficient $\frac{1}{\sqrt{\text{SNR($t$)}}}$ is expected to be significantly larger as the step $t$ increases (as shown in Fig.~\ref{fig:main}), which would result in a substantial deviation of the estimated $\hat{x}_{0}$ from the target $x_0$. 
We also visualize the estimated $\hat{x}_0$ and the amplified error at different timesteps via feeding $x_t=\sqrt{\alpha _{t}}x_{0} + \sqrt{1-\alpha _{t}}\epsilon$ into the denoising network once. The $\mathit{estimated \ \hat{x}_0}$ increasingly deviates from the ground-truth $x_0$ when $t$ grows larger, meanwhile the $\mathit{amplified \ error}$ becomes larger and even gradually approaches $x_0$. In this regard, we can find that the constant training weight is indeed biased, and optimizing the explicit target $\epsilon$ uniformly across different timesteps cannot guarantee approaching the implicit target $x_0$ exactly.

\subsection{Improved Training Strategy}
The above theoretical analysis provides a principled guidance on coping with the biased estimation problem and designing the loss weighting strategy (existing weighting strategies can be covered from unified perspective under the proposed principle in Subsec. \ref{subsec:unified}). Concretely, besides expecting the loss function to reach the explicit target $\epsilon$, which is relatively simple, more importantly, we desire to encourage the estimated $\hat{x}_0$ to approach the implicit target $x_0$. Therefore, it is essential to consider the varying impact of noise prediction at different steps $t$ when designing the loss weight. In this regard, we adopt the amplification coefficient in the amplified error part of Eq. \ref{eq:8} as the loss weighting coefficient:
\begin{equation}
\label{eq:11}
    L=\sum _t \mathbb{E}_{x_0,\epsilon}\left[\frac{1}{\sqrt{SNR(t)}}||\epsilon-\epsilon_\theta(x_t, t)||^2\right].
\end{equation}

In other word, we assign higher weight as the step $t$ increases (i.e., when adding more noise to $x_0$), thereby compelling the noise error $(\epsilon_\theta(x_t, t) - \epsilon)$ to decrease more significantly at larger step $t$.
Note that the key of this paper is the comprehensive exploration of the bias issue in diffusion models. Grounded in our unlocked bias problem, a simple loss weight design can still achieve substantial performance improvement (refer to the analyses and experiments in the following sections). Besides, we also provide more discussions on the weight selection in the Sec. \ref{subsec:unified} and the supplementary material.

\section{Comprehensive Understanding the Bias Problem}
In this section, we aim to address several key questions crucial for achieving a systematical understanding of the bias problem in conventional diffusion models: Why is the bias problem important? What are its effects? And what is the underlying cause? We believe answering these questions is essential for unraveling the black box of diffusion models.

\subsection{Biased Estimation in Training Process}   \label{section:existence}
\begin{figure}[t]
	\begin{center}
		\includegraphics[width=0.98\linewidth]{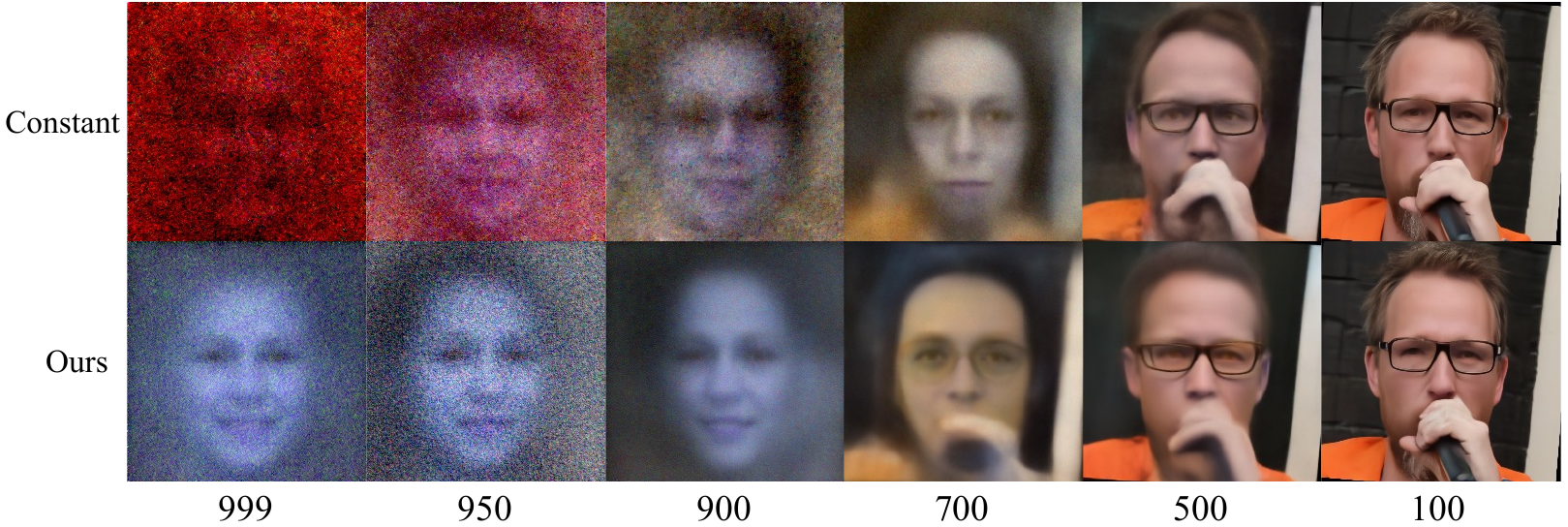}
	\end{center}
        \setlength{\abovecaptionskip}{-0.2cm}
        \setlength{\belowcaptionskip}{-0.4cm}
	\caption{We present the one-step estimation results of $\hat{x}_0$ using different input samples $x_t$, where the diffusion models are pretrained on the FFHQ dataset \cite{karras2019style} with different loss weighting strategies. One-step estimation: start from a clean image and add noise to get $x_t$ according to Eq.~\ref{eq:3}. Then put $x_t$ into the denoising network once to get the estimated noise $\hat{\epsilon}$, and the corresponding $\hat{x}_0$. The top row displays the results obtained using a well-trained constant weighting model, while the bottom row depicts the results achieved with our well-trained improved weighting model.}
	\label{fig:error}
\end{figure}

First, 
we illustrate the one-step estimation $\hat{x}_0$ in Fig.~\ref{fig:error} to compare the results obtained using the original constant weighting and our variant. There is a general tendency for the estimated $\hat{x}_0$ of both weighting strategies to gradually deviate from the original $x_0$ as the step $t$ increases, which is inevitable due to the increasing noise in the input $x_t$. However, when utilizing the constant weighting loss for training, noticeable color shifts and inferior arrangement of human faces can be observed in the early steps ($t=999$ and $t=950$), severely deviating from the target $x_0$. In contrast, our strategy effectively reduces the bias, achieving greater consistency with the targets across various timesteps, even under relatively high noise levels (e.g., at $t=999$ and $t=950$). These findings indicate that the proposed weighting strategy facilitates training in a more appropriate direction. 
More analyses are available in Sec. \ref{ap:existence}.

\subsection{Biased Generation on Sampling Process}  \label{section:impact}
We further analyse the detrimental effects of the biased estimation problem introduced by the constant weighting loss for model inference, i.e., \textit{biased generation} on the sampling process. As seen in the first two rows in Fig.~\ref{fig:fewer_steps}, biased generation primarily attributes to the chaos and inconsistency in the early few sampling steps, which substantially affects the final generation through error propagation.
We particularly observe pronounced color shifting in biased generation when employing a small number of sampling steps (e.g., $T=2$), which remains challenging to correct even with an extended sampling process (e.g., $T=1000$). In contrast, training with our strategy can essentially prevent the issue (e.g., the shown images with $T=2$), eliminating the need for a lengthy correction process. Moreover, generated images using our strategy show enhanced details and global consistency compared to the baseline method. More visual results and analyses are available in supplementary material.

\subsection{Underlying Causes of Biased Estimation}
\label{section:reason}
Finally, we take one step further to unravel the underlying causes of biased estimation. Specifically, we find that the optimization difficulty and importance of the denoising network is vastly different across step $t$.

\begin{figure}[ht]
	\begin{center}
		\includegraphics[width=0.98\linewidth]{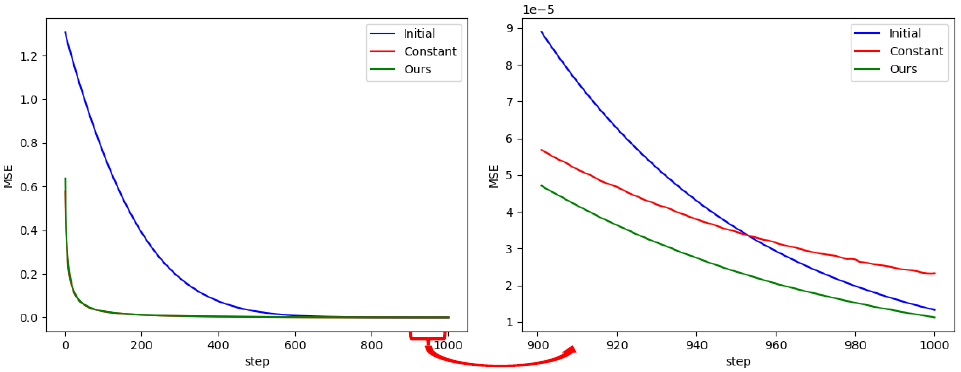}
	\end{center}
        \setlength{\abovecaptionskip}{-0.2cm}
        \setlength{\belowcaptionskip}{-0.4cm}
	\caption{MSE-step curve under several settings. ``Initial'' mode is calculated between input and target. Obviously, the optimization difficulty is vastly different across step $t$. ``Constant'' and ``Ours'' modes are calculated between network output and target, and ``Constant'' denotes constant weight strategy and ``Ours'' stands for our proposed weight strategy. \textbf{Note that the green and red curve visually overlap in the left figure due to large scale.}}
	\label{fig:MSE}
\end{figure}

\noindent \textbf{Different optimization difficulty.}
Intuitively, the input $x_t$ is closer to the target as step $t$ becomes larger. Consequently, the network encounters varying levels of fitting difficulty across different values of $t$, with larger values of $t$ being relatively easier. To verify this, we plot the Mean squared error (MSE)-step curve under several settings in Fig.~\ref{fig:MSE}. In the ``Initial'' setting, the MSE value is directly computed between the network input $x_t$ and the target Gaussian noise. The remaining two settings compute the MSE value between the network output and the target Gaussian noise, with ``Constant'' representing the constant weighting method and ``Ours'' representing the proposed weighting strategy. The distribution of MSE value under ``Initial'' mode is extremely unbalanced, in which the MSE value is negligible when $t > 600$. Consequently, this imbalance endows different optimization difficulty across step $t$ and renders the constant weighting strategy suboptimal. Specifically, when $t$ is sufficiently large, the MSE value between the network input and the target becomes extremely small, allowing the network to ``do nothing'' while still maintaining a low MSE loss.

The above analysis is verified in the right part of Fig.~\ref{fig:MSE}. For $t > 950$, the MSE value in constant weight mode surpasses that of the ``Initial'' mode, indicating the output deviates even further from the target than the input. \textbf{This observation illustrates that the denoising network in constant weight setting fails to identify the noise pattern in the input at large step $t$ and, therefore, cannot effectively handle the denoising task.} In contrast, our weight strategy consistently yields MSE values lower than those of the ``Initial'' mode, demonstrating its exceptional denoising capability, particularly for highly noisy inputs.

\noindent \textbf{Different importance.}
Lastly, we reveal that the importance of the denoising network also varies across step $t$.
Intuitively, initial steps are important for both training and sampling process. For training, the initial steps pose greater difficulty due to high noise levels in the input. For sampling, the initial steps serve as the foundation for subsequent steps, contributing to error propagation.
Theoretically, we have verified that initial steps should be emphasized to reach the implicit target $x_0$ in Sec. \ref{section:theoretical}. Additionally, we also find evidence supporting the crucial role of initial steps in diffusion models \cite{nichol2021improved, wang2023boosting}. For example, Nichol et al. \cite{nichol2021improved} demonstrated that the first few steps contribute the most to the variational lower bound. Wang et al. \cite{wang2023boosting} found that reusing update directions from initial steps with adaptive momentum sampler can generate images with enhanced details. The constant weighting strategy assumes equal importance across all steps. While, our method assigns higher weights to the initial steps, which is consistent with both intuition and theory.

\section{Experiments}

\subsection{Setup} \label{sec:setup}
\noindent \textbf{Datasets.}
We perform experiments on unconditional image generation using the FFHQ \cite{karras2019style}, CelebA-HQ \cite{karras2017progressive}, AFHQ-dog \cite{choi2020stargan}, and MetFaces \cite{karras2020training} datasets. These datasets contain approximately 70k, 30K, 50k, and 1k images respectively. Besides, we conduct class-conditional generation on CIFAR-10 \cite{krizhevsky2009learning} and ImageNet \cite{deng2009imagenet} datasets. 
We resize and center-crop data to 256×256, following the pre-processing performed by ADM \cite{dhariwal2021diffusion}.

\noindent \textbf{Training details.}
We set T = 1000 for all experiments. We implement the proposed approach on top of ADM \cite{dhariwal2021diffusion}, which offers well-designed architecture. We train our model for 500K iterations with a batch size of 8.

\noindent \textbf{Evaluation settings.}
Following the common practice \cite{song2020improved}, we utilize an Exponential Moving Average (EMA) model with a rate of 0.9999 for all experiments. Besides, we generate 50K samples for each trained model and use the full training set to compute the reference distribution statistics, following \cite{ho2020denoising, choi2022perception}. During inference, we obtain results with fewer sampling steps than T by employing the respacing technique.
For quantitative evaluations, we employ the Fréchet Inception Distance (FID) \cite{heusel2017gans}.

\subsection{Comparison to Existing Weighting Strategies}  \label{subsec:unified}

\begin{figure*}[ht]
	\begin{center}
		\includegraphics[width=0.98\linewidth]{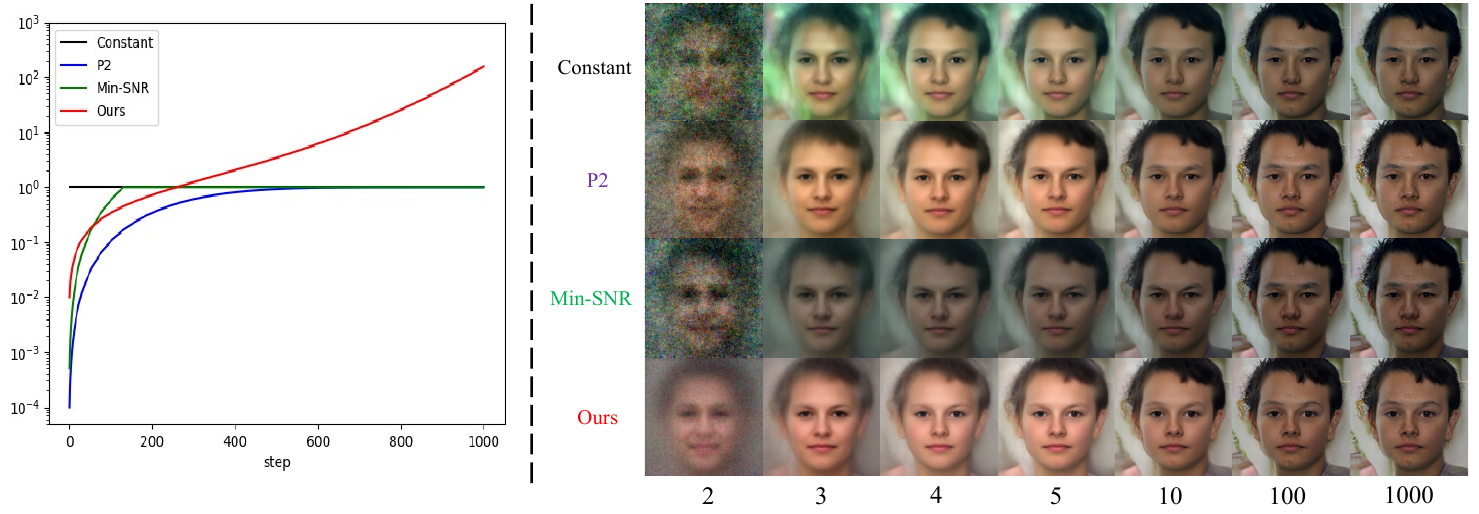}
	\end{center}
        \setlength{\abovecaptionskip}{-0.2cm}
        \setlength{\belowcaptionskip}{-0.4cm}
	\caption{Left: Visualization of various weighting strategies. P2 and Min-SNR start from the basis of constant weight and lower the weight down for small $t$. Right: Sampling results with different total sampling steps $T$. From top to bottom, they are constant, P2, Min-SNR, and our method. Evidently, P2 and Min-SNR still suffer from bias and artifacts during the initial generation stage.}
	\label{fig:weight}
\end{figure*}

\noindent \textbf{Unified perspective on existing weighting strategies.}
Some methods explore various weighting strategies in $\epsilon$-prediction mode, including P2 \cite{choi2022perception} and Min-SNR \cite{hang2023efficient}. We present these distinct weighting strategies in Fig.~\ref{fig:weight}.
Most methods modify the weight on the basis of the conventional constant weighting with intuition or observation. Compared to using constant weights, they only lower the weights for small $t$, maintaining the weights unchanged for the remaining substantial portion of the steps. Besides, they encounter difficulties in establishing general principles for guiding the design of the weighting strategy. 
In contrast, we can take a unified perspective on these weighting strategies within the framework of the unlocked bias analyses. Specifically, the amplification coefficient in Eq. \ref{eq:8} serves as a general principle on the loss weight design. Our theoretically principle elucidates that the weight should monotonically increase as $t$ increases, as depicted with the red curve in Fig. \ref{fig:weight}. Prior methods \cite{choi2022perception, hang2023efficient} tend to assign lower weights to small $t$ values, adhering to the principle overall. These observations can also substantiate the rationale behind their superior performance in comparison to constant weighting. We can also demonstrate that using our improved formulation can further enhance performance, leveraging insights gained from the analysis of bias issues.

The related works in \cite{salimans2021progressive, karras2022elucidating} employ different training objectives, and detailed experimental comparison and discussions on these works can be found in the supplementary material.

\begin{table}[t]
\small
\setlength{\tabcolsep}{8pt}
\caption{Quantitative comparison on unconditional generation. The experimental results are reported in terms of FID under a fair setting, with the only distinction being the loss weighting strategy. * denotes the results reported in the original paper. However, as certain essential training details of P2* (e.g., training iterations) are unknown, its reported values are used for reference only.} 
\label{table:performance_Comparison}
\centering
\setlength{\abovecaptionskip}{-0.2cm}
\setlength{\belowcaptionskip}{-0.2cm}
\begin{tabular}{lrccccc}
	\toprule
        Dataset & Step $T$ & Constant & P2 & P2* & Min-SNR & Ours \\
        \midrule
	\multirow{6}{*}{FFHQ} & 1000  & 10.864 & 6.517  & 6.92 & 6.501  & 6.354    \\
                              & 500   & 11.027 & 6.792  & 6.97 & 6.873  & 6.706    \\
                              & 250   & 11.780 & 7.478  & -    & 7.722  & 7.385    \\
                              & 100   & 15.671 & 10.855 & -    & 11.391 & 10.815   \\
                              & 50    & 22.375 & 16.538 & -    & 17.328 & 15.345   \\
                              & 20    & 41.270 & 34.399 & -    & 34.652 & 29.380   \\
 	\midrule              
	\multirow{2}{*}{CelebA-HQ} & 1000  & 9.374  & 7.258  & 6.91 & 6.322  & 5.980    \\
                                   & 500   & 10.236 & 7.718  &      & 6.923  & 6.572    \\
                                   & 250   & 11.097 & 8.433  & -    & 8.016  & 7.604    \\
                                   & 100   & 12.006 & 9.297  & -    & 9.385  & 8.836    \\
 	\midrule
	\multirow{4}{*}{AFHQ-dog}  & 1000  & 18.300 & 17.068  & 11.55 & 17.342 & 14.928     \\
                                   & 500   & 18.606 & 17.474  & -     & 17.639 & 14.946     \\
                                   & 250   & 19.104 & 17.759  & 11.66 & 17.922 & 15.033     \\  
                                   & 100   & 20.446 & 18.344  & -     & 18.421 & 15.821     \\
 	\midrule
	\multirow{4}{*}{MetFaces}& 1000  & 41.418 & 14.204  & -     & 30.876 & 9.168    \\
                                 & 500   & 42.115 & 14.448  & -     & 31.168 & 9.429    \\
                                 & 250   & 42.324 & 14.738  & 36.80 & 31.340 & 9.849    \\
                                 & 100   & 42.624 & 14.994  & -     & 31.626 & 10.388   \\
	\bottomrule
\end{tabular}
\end{table}

\begin{table}[ht]
\small
\setlength{\tabcolsep}{8pt}
\caption{Quantitative comparison on class-conditional generation. We employ FID metric to evaluate the distribution distance and IS metric to evaluate the sampling diversity. Our method achieves better results on both metrics.} 
\label{table:performance_Comparison_class}
\centering
\setlength{\abovecaptionskip}{-0.2cm}
\setlength{\belowcaptionskip}{-0.2cm}
\begin{tabular}{lrccccc}
	\toprule
        Dataset & Step $T$ & Metrics & Constant & P2 & Min-SNR & Ours \\
        \midrule
	\multirow{2}{*}{CIFAR-10} & \multirow{2}{*}{1000}  & FID $\downarrow$ & 11.97 & 11.71 & 9.02 & 8.45    \\
                                                     &      & IS $\uparrow$   & 8.07 & 8.11 & 8.14 & 8.13    \\
 	\midrule
	\multirow{2}{*}{ImageNet} & \multirow{2}{*}{100}   & FID $\downarrow$ & 99.00 & 95.47 & 94.81 & 93.32    \\
                                                      &     & IS $\uparrow$   & 11.96 & 13.02 & 13.21 & 13.37    \\
	\bottomrule
\end{tabular}
\end{table}

\noindent \textbf{Quantitative comparison.}
Tab. \ref{table:performance_Comparison} presents a quantitative performance comparison of various weighting strategies across different sampling steps $T$. Our method substantially lifts the performance limit across multiple datasets and sampling steps. These datasets are of various scales ranging from 1k to 70k images, which indicates the generalization and robustness of our method. Besides, our method can effectively elevate the performance of the constant weight baseline on all the possible total sampling steps. It is worth noting that the performance gain is particularly pronounced with smaller datasets and shorter sampling steps, which matches our theoretical derivation and extensive analyses. 

Tab. \ref{table:performance_Comparison_class} presents a quantitative performance comparison on class-conditional generation of various weighting strategies. Besides FID, we also adopt Inception Score (IS) to measure the generation diversity. Obviously, our method surpasses previous methods with better FID score and higher diversity.

\begin{figure*}[t]
	\begin{center}
		\includegraphics[width=0.95\linewidth]{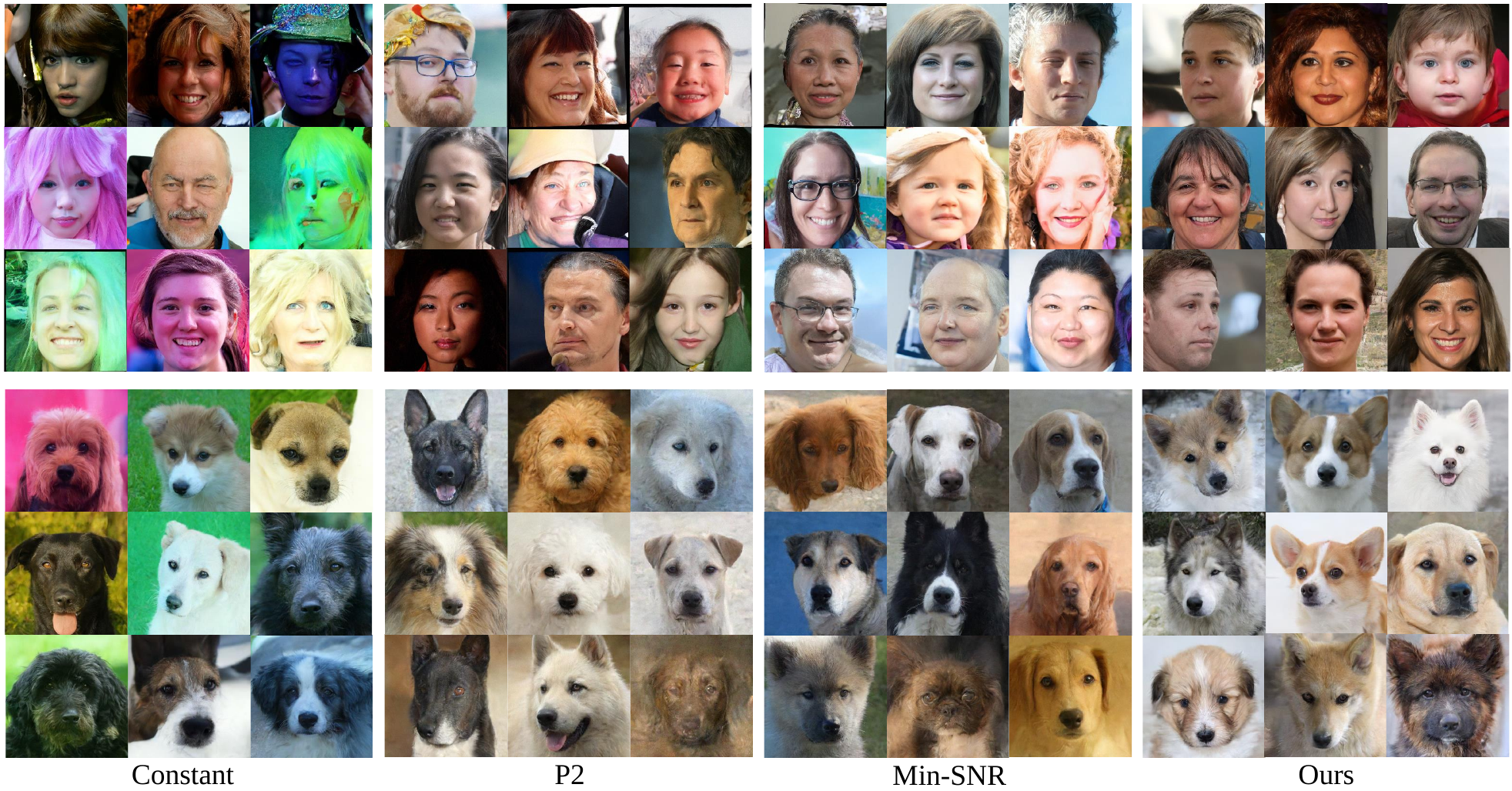}
	\end{center}
        \setlength{\abovecaptionskip}{-0.2cm}
        \setlength{\belowcaptionskip}{-0.2cm}
	\caption{Visual results of different weighting strategies on different datasets. We randomly choose the first nine generated images without cherry-pick. The first row is trained on FFHQ dataset and the second row is on AFHQ-dog dataset.}
	\label{fig:samples}
\end{figure*}

\noindent \textbf{Qualitative comparison.}
Fig.~\ref{fig:samples} presents the qualitative results. As anticipated, the biased constant weighting strategy produces images with inferior global structure and color alignment. P2 and Min-SNR enhance the sample quality by building upon the constant weight foundation. However, they still produce images with inferior global structure. This is due to their significant bias and chaotic behavior during the initial sampling steps, as depicted in Fig.~\ref{fig:weight}. In contrast, our method is totally free of the dilemma of color shift.

\begin{figure}[ht]
	\begin{center}
		\includegraphics[width=0.98\linewidth]{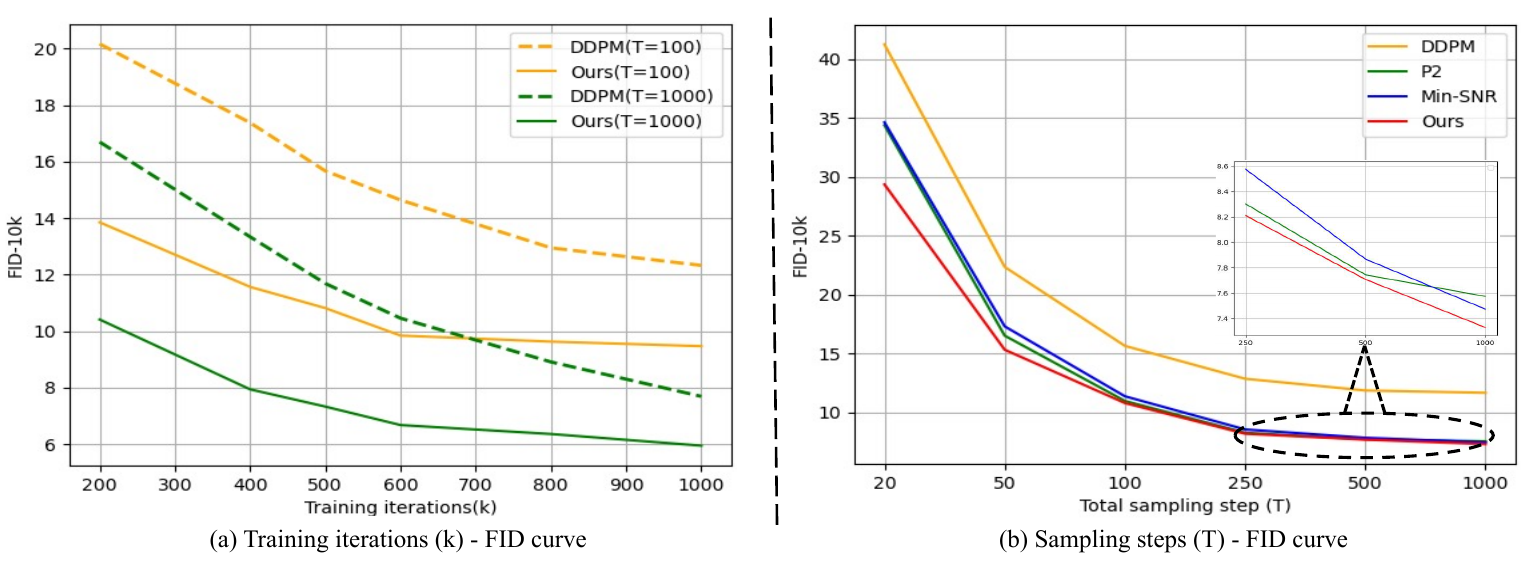}
	\end{center}
        \setlength{\abovecaptionskip}{-0.2cm}
        \setlength{\belowcaptionskip}{-0.2cm}
	\caption{(a): FID-training iterations curve. (b): FID-sampling steps curve. These two curves are obtained on FFHQ dataset. Our method is more efficient and high-performing. Note that, we use DDPM to denote the constant weighting strategy.}
	\label{fig:FID_trainsample}
\end{figure}

\subsection{More Analyses}  \label{ap:existence}
\noindent \textbf{High efficiency.}
Fig.~\ref{fig:FID_trainsample} illustrates the FID-training iterations curve and the FID-sampling steps curve for the FFHQ dataset. The training curve clearly demonstrates the superior efficiency and potential of our method. For instance, our weighting strategy matches the performance of 1000k iterations of constant weight training with only 400k iterations. In terms of sampling, our method surpasses all existing weight strategies across all sampling steps. Moreover, consistent with the analysis in Sec. \ref{section:impact}, the performance gains are more pronounced with fewer sampling steps.

\begin{figure*}[ht]
	\begin{center}
		\includegraphics[width=0.98\linewidth]{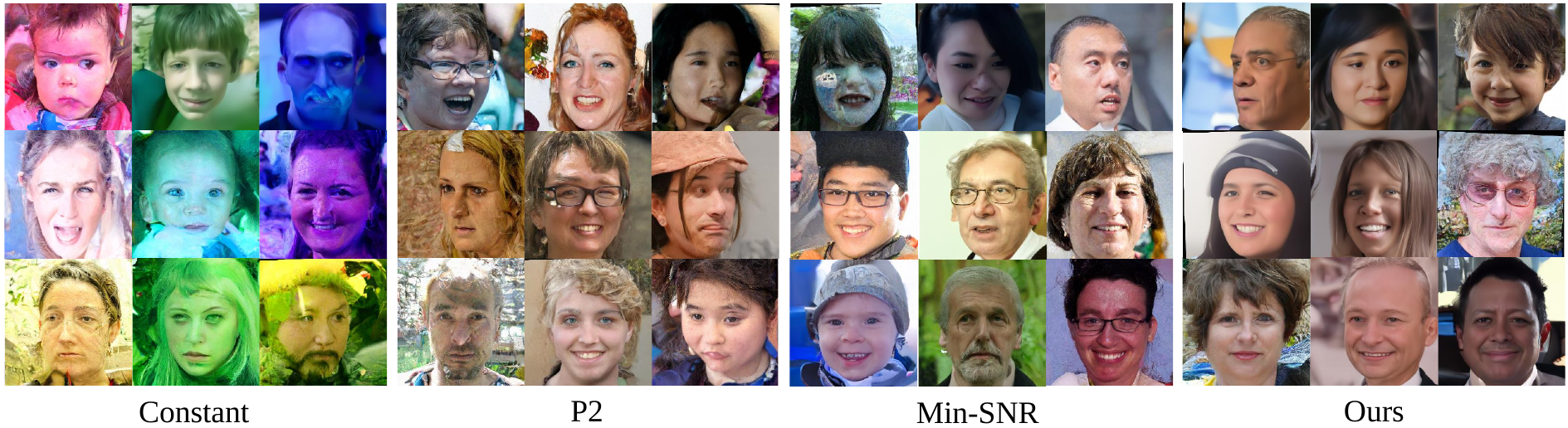}
	\end{center}
        \setlength{\abovecaptionskip}{-0.2cm}
        \setlength{\belowcaptionskip}{-0.2cm}
	\caption{The generated samples of DDIM sampler under four different weighting strategies on FFHQ dataset.}
	\label{fig:DDIM}
\end{figure*}

\begin{figure}[ht]
	\begin{center}
		\includegraphics[width=0.98\linewidth]{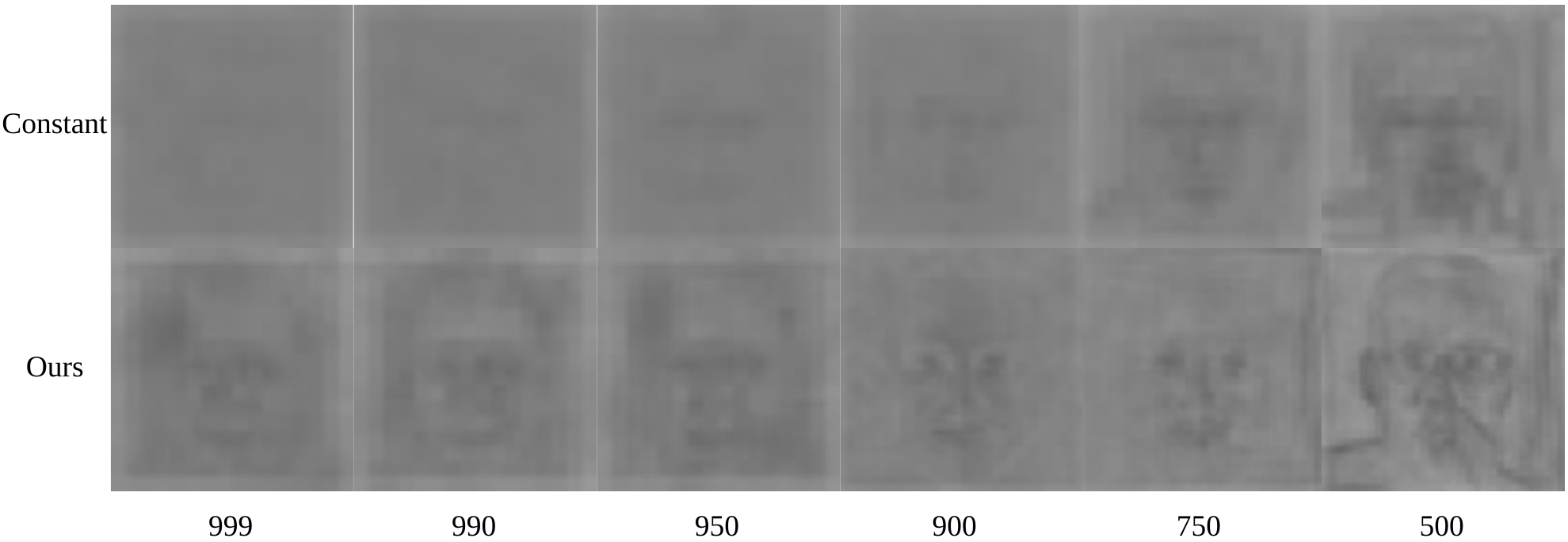}
	\end{center}
        \setlength{\abovecaptionskip}{-0.2cm}
        \setlength{\belowcaptionskip}{-0.2cm}
	\caption{The intermediate feature maps at different steps $t$. Intermediate feature maps are correlated with image structure underlying the noisy samples. Obviously, constant strategy struggles to generate clear facial architecture with noisy $x_t$ as input ($t>$900). In contrast, our method can generate clear facial layout even with the most noisy $x_{999}$ as input.}
	\label{fig:existence}
\end{figure}
\noindent \textbf{Different samplers.}
Our weighting strategy is orthogonal to samplers. We conduct additional DDIM sampler \cite{song2020denoising} to validate this conclusion. As shown in Fig.~\ref{fig:DDIM}, we depict the generated samples of DDIM sampler under four different weighting strategies on FFHQ dataset. Similar to the conclusion in DDPM sampler in Fig.~\ref{fig:samples}, our method achieves the highest performance with DDIM sampler among these four weighting strategies.

\noindent \textbf{More analyses of the bias in training Process.}
In this part, we give more analyses and visualization to validate the existence of the bias problem in the training process. Previous image editing methods \cite{hertz2022prompt, tumanyan2023plug} find that the intermediate feature maps can reflect the structures underlying the noisy samples. We follow this practice with the intuition that the bias problem may also lead to poor structure generation. Concretely, we present the intermediate feature maps at various step $t$ in Fig.~\ref{fig:existence}. Consistent with the conclusion in Fig.~\ref{fig:error}, the constant weight mode generates poor structures with relatively large noise scale. For example, it struggles to generate clear facial structure when step $t>$500 (nearly half of the range field). On the contrary, our method has clear facial layout 
across all timesteps. Especially, the facial structure is visible with the most noisy $x_{999}$.

\noindent \textbf{Comparison to prior literature.}
Performance comparison between our method and existing generative methods on the FFHQ dataset \cite{karras2019style} is presented in Tab. \ref{table:performancen}. Previous methods \cite{brock2018large, karras2019style, sauer2021projected, esser2021taming, rombach2022high, ho2022cascaded, bao2023all} achieve exceptional results with meticulously designed architectures and methodologies. In contrast, our method achieves competitive performance with a simple loss weight strategy. Besides, our method is a general strategy for diffusion models and can further elevates their performance. For instance, we achieved substantial improvements by solely adjusting the loss weight on top of ADM \cite{dhariwal2021diffusion}, reducing the FID score from 10.86 to 6.35.
Moreover, our method offers the capability to achieve even higher performance. Firstly, we can extend the training duration. For instance, with 500k iterations, our method achieves a FID of 6.35, while with 1000k iterations, it achieves a FID of 4.97. Additionally, we have the flexibility to replace the codebase ADM with a stronger model, such as stable diffusion.

\begin{table}[ht]
\small
\setlength{\tabcolsep}{6pt}
\caption{Quantitative comparison to prior generative models on FFHQ dataset. Our method is on top of ADM with only one additional line of code, yet achieving substantial performance lift.}
\label{table:performancen}
\centering
\begin{tabular}{lllr}
	\toprule
        Dataset & Method & Type & FID \\
        \midrule
	\multirow{8}{*}{FFHQ} & BigGAN \cite{brock2018large}         & GAN & 12.4    \\
                              & UNet GAN \cite{schonfeld2020u}       & GAN & 10.9    \\
                              & StyleGAN \cite{karras2019style}      & GAN & 4.16    \\
                              & StyleGAN2 \cite{karras2020analyzing} & GAN & 3.73    \\
                              & VQGAN \cite{esser2021taming}         & GAN+AR & 9.6   \\
                              & LDM \cite{rombach2022high}           & Diffusion model & 4.98   \\
                              & ADM (Baseline) \cite{dhariwal2021diffusion}  & Diffusion model & 10.86   \\
                              & Ours (500k iterations)                                 & Diffusion model & 6.35   \\
                              & Ours (1000k iterations)                                   & Diffusion model & 4.97   \\
	\bottomrule
\end{tabular}
\end{table}

\subsection{Discussions}
We unlock the biased training problem of diffusion models, which lies as the key of our method. Grounded in our unlocked bias problem, the proposed simple loss weight design can achieve substantial performance improvement.
Given that diffusion models usually serve as fundamental building blocks for various application-oriented works, our method provides valuable inspiration and insights for these endeavors.
Additionally, we identify several potential avenues for future research. (1) The elucidated mechanism behind the biased problem offers valuable insights for downstream tasks, such as editing and restoration, facilitating the integration of the bias issue into specific tasks. (2) The biased problem can be investigated from other perspectives, such as noise schedule \cite{ning2023input, chen2023importance}. It is encouraging to discuss the defects of diffusion models from a unified perspective. 

\section{Conclusion}
This paper provides theoretical analyses and comprehensive studies to demonstrate that the traditional uniform weighting loss function is suboptimal, by examining the existence, impact, and underlying reasons behind this issue. To mitigate this problem, we employ a simple yet highly effective weighting strategy. Empirical studies conducted on multiple datasets, along with comparisons with existing weight methods, further validate the effectiveness of our approach. We also believe these analyses contribute to a deeper understanding of the underlying mechanism of diffusion models.

\paragraph
\noindent \textbf{Acknowledgements.}
This work was supported by the Anhui Provincial Natural Science Foundation under Grant 2108085UD12. We acknowledge the support of GPU cluster built by MCC Lab of Information Science and Technology Institution, USTC.

\bibliographystyle{splncs04}
\bibliography{main}



\newpage
\appendix
\onecolumn

\begin{center}
    \Large{\textbf{Appendix}}
\end{center}

\section{Full Derivation of the Training Objectives}  \label{ap:derivation}
Diffusion models are trained by optimizing a variational lower bound (VLB). For each step $t$, the denoising score matching loss $L_t$ is the distance between two Gaussian distributions, which can be rewritten as:
\begin{align}
\label{eq:1_sup}
    L_{t}=D_{KL}(q(x_{t-1}|x_t,x_0)~||~p_\theta(x_{t-1}|x_t)),
\end{align}
where the reverse diffusion step $q(x_{t-1}|x_t,x_0)$ and $p_\theta(x_{t-1}|x_t)$ can be expressed as follows:
\begin{equation}
\label{eq:2_sup}
    \begin{aligned}
    & q(x_{t-1}|x_t,x_0) = N(x_{t-1};\tilde{\mu}_t(x_t,x_0),\tilde{\beta}_t \mathbf{I}), \\
    & p_\theta(x_{t-1}|x_t) = N(x_{t-1};\mu _\theta(x_t,t),{\textstyle \sum_{\theta}^{}} (x_t,t)), \\
    \end{aligned}
\end{equation}
where $\tilde{\mu}_t(x_t,x_0) := \frac{\sqrt{\alpha_{t-1}}\beta_t}{1-\alpha_t}x_0 +  \frac{\sqrt{1 - \beta_t}(1-\alpha_{t-1})}{1-\alpha_t}x_t$,  $\tilde{\beta}_t:= \frac{1-\alpha_{t-1}}{1-\alpha_t}\beta_t$, and the variance ${\textstyle \sum_{\theta}^{}} (x_t,t)=\sigma_t^2 \mathbf{I}$. Ho et al. \cite{ho2020denoising} set $\sigma_t^2 = \beta_t$. Thus, we can rewrite $L_{t}$ as follows:

\begin{equation}
\label{eq:3_sup}
    \begin{aligned}
    L_{t}&=\mathbb{E}_{x_0,\epsilon}\left[\frac{1}{2 \sigma_{t}^{2}}\left\|{C_1}x_0 +  {C_2} x_t  -\boldsymbol{\mu}_{\theta}\left(\mathbf{x}_{t}, t\right)\right\|^{2}\right] + C, \\
    & C_1 = \frac{\sqrt{\alpha_{t-1}}\beta_t}{1-\alpha_t},   C_2 = \frac{\sqrt{1 - \beta_t}(1-\alpha_{t-1})}{1-\alpha_t} \\
    \end{aligned}
\end{equation}

Thus, the denosie network $\mu_{\theta}(x_t, t)$ can be optimized to predict $x_0$. Further, $x_0$ deterministically corresponds to $\epsilon$ as:
\begin{equation}
\label{eq:4_sup}
    x_0=\frac{1}{\sqrt{\alpha _{t}}} x_{t} - \sqrt{\frac{1-\alpha _{t}}{\alpha _{t}} }\epsilon,
\end{equation}
Thus, we can also set the training target to be $\epsilon$ via replacing the $x_0$ in Eq. \ref{eq:3_sup} with Eq. \ref{eq:4_sup} to get:
\begin{equation}
\label{eq:5_sup}
    L_{t}=\mathbb{E}_{x_0,\epsilon}[\frac{\beta_t^2}{(1-\beta _t)(1-\alpha _t)}||\epsilon-\epsilon_\theta(x_t, t)||^2].
\end{equation}

\section{More Discussions on the Weight}  \label{ap:optimal weight}
We treat $\epsilon$ as the explicit target and $x_0$ as the implicit target. Therefore, $\frac{1}{SNR(t)}$ weight seems reasonable for optimizing $x_0$ but hinders the optimization of $\epsilon$. Constant weight seems reasonable for optimizing $\epsilon$ but is biased from $x_0$. By taking both targets and the bias formulation into account, we employ the amplifying factor as the loss weight to form the loss strategy in Eq. 9 of the main manuscript, which is basically consistent with the the amplified error part in Eq.8 of the main manuscript. Such elegant loss weight design is theoretically feasible based on Eq.8 and can help minimize the the amplified error part with experimental verification. 

\begin{figure*}[ht]
	\begin{center}
		\includegraphics[width=0.98\linewidth]{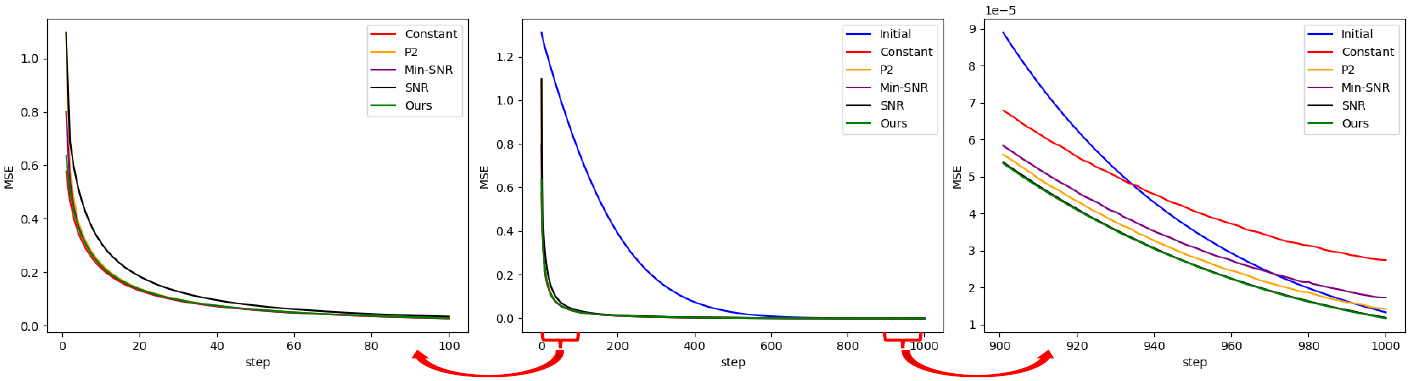}
	\end{center}
        \vspace{-0.2cm}
	\caption{MSE analysis of different weighting strategies. We use ``SNR'' to denote the weight of $\frac{1}{SNR}$. SNR weight can't further lower the MSE for large $t$. On the contrary, it's denoising ability is worsen for small $t$ with large MSE. The reason behind this is the excessive range field of SNR weight ranging from $10^{-4}$ to  $10^{4}$, which causes the denoising network excessively focusing on few early steps. For example, for batchsize=8, if one $t$ is large and the remaining seven $t$ are small, the network will pay excessive attention to the large $t$ with high weight, while at the cost of sacrificing the seven small $t$s. In contrast, our method achieves the lowest MSE across these weighting strategies at different step $t$, only slightly larger than the constant weight for $t < 50$.}
	\label{fig:snr}
\end{figure*}

\begin{figure}[ht]
	\begin{center}
		\includegraphics[width=0.98\linewidth]{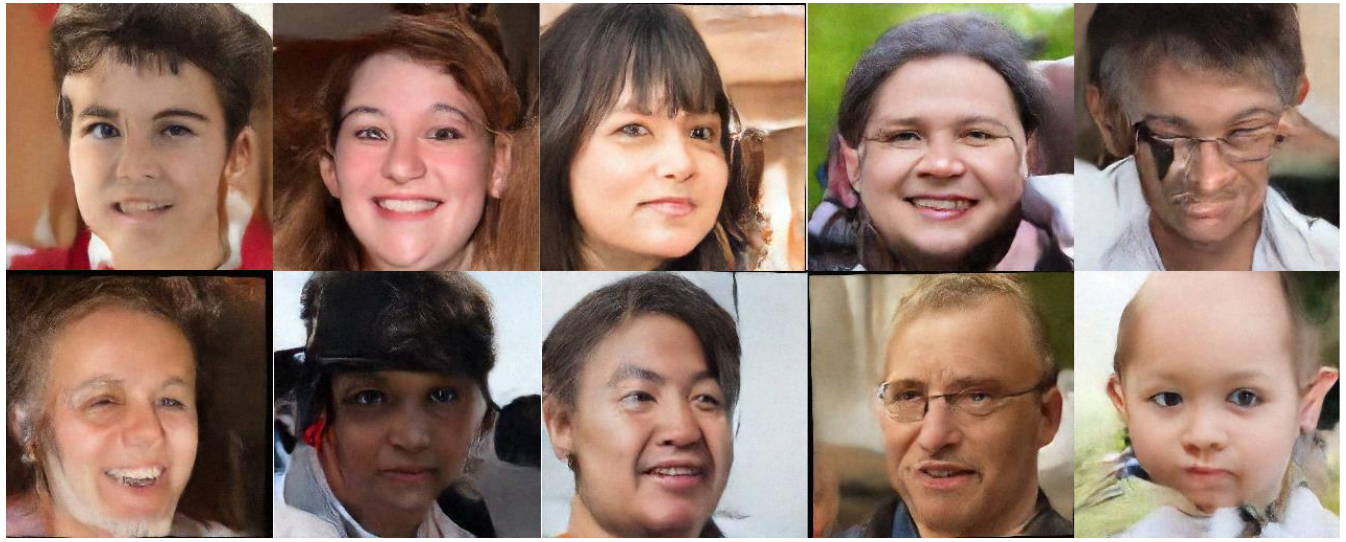}
	\end{center}
        \vspace{-0.2cm}
	\caption{First ten generated sample of ''SNR'' weight. This weight strategy leads to divergence and poor sample quality.}
	\label{fig:SNRsample}
\end{figure}

Besides, we further discuss the $\frac{1}{SNR(t)}$ weight. We use ``SNR'' to denote the weight of $\frac{1}{SNR}$. We demonstrate its sub-optimality from two aspects. (1) SNR weight damages the explicit target without further boosting the implicit target as shown in Fig.~\ref{fig:snr}. Concretely, the MSE of the SNR mode completely overlaps with our $\frac{1}{\sqrt{SNR}}$ weigh strategy for large $t$, indicating that SNR weight can't further boost the implicit target. However, the MSE of the SNR mode is substantially larger than all other weighting strategies for small $t$, indicating that the explicit target is seriously violated. The reason behind this is the excessive range field of SNR weight ranging from $10^{-4}$ to  $10^{4}$, which causes the denoising network excessively focusing on few early steps. (2) Empirically, the SNR mode performs terribly, as shown in Fig.~\ref{fig:SNRsample}. Min-SNR \cite{hang2023efficient} also explores predicting Gaussian noise with the weight of $\frac{1}{SNR}$, and they find that this setting leads to divergence. Thus, our experimental result is also consistent with the conclusion of Min-SNR.

\section{More Analyses of the Biased Generation}  \label{ap:biased generation}
\begin{figure}[ht]
	\begin{center}
		\includegraphics[width=0.98\linewidth]{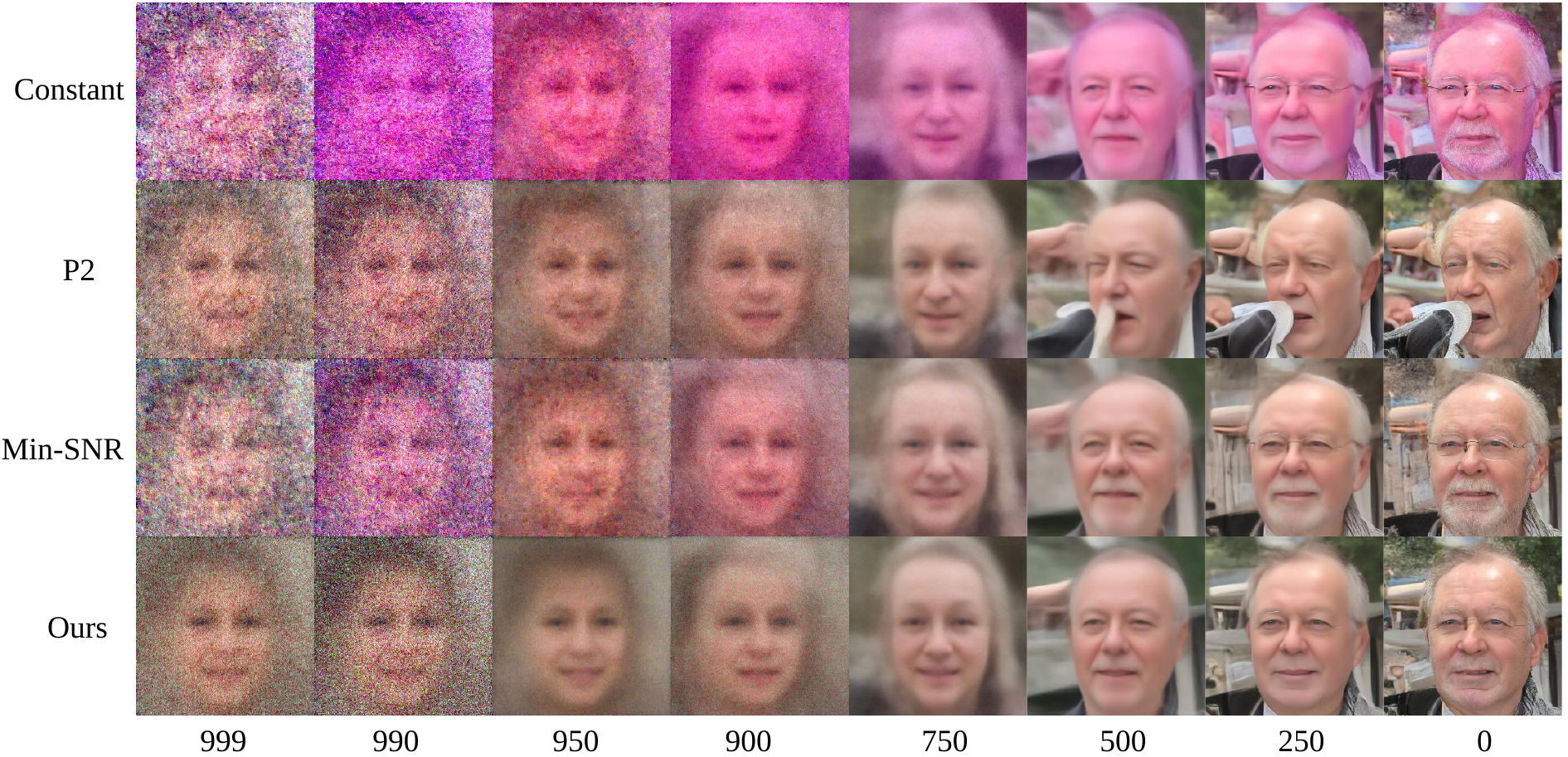}
	\end{center}
	\caption{Biased generation: the estimated $\hat{x}_0$ of different weighting strategies at different step $t$. Constant weight exhibits artifacts and color shift in the first generation step, resulting in the final generated images with color and structure distortion. P2 and Min-SNR also show global artifact and inconsistency in the first generation step. Thus, their generated images suffer from poor structures.}
	\label{fig:one_step}
\end{figure}

\begin{figure}[ht]
	\begin{center}
		\includegraphics[width=0.98\linewidth]{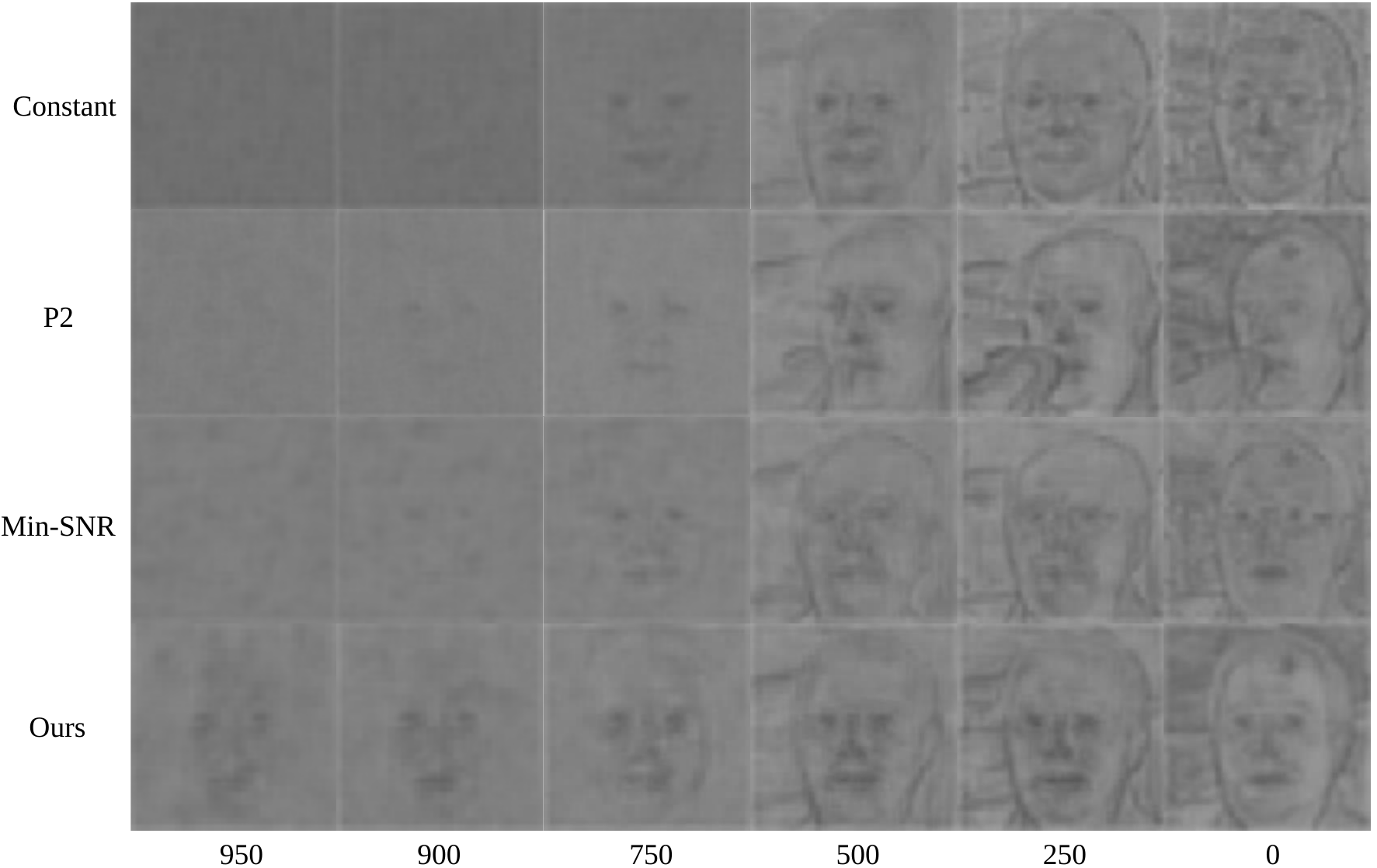}
	\end{center}
	\caption{Biased generation: the intermediate feature maps of different weighting strategies at different steps $t$. Constant, P2, and Min-SNR weight strategies struggle to generate clear facial architecture in the early generation steps. Our method shows clear facial architecture even at very early steps.}
	\label{fig:features}
\end{figure}

In the main manuscript, we indicate the biased generation with generated samples of different total sampling steps $T$. In this section, we show more analyses and visualization of the biased generation. As shown in Fig.~\ref{fig:one_step} and Fig.~\ref{fig:features}, we respectively show the estimated $\hat{x}_0$ and intermediate feature maps of different weighting strategies at different step $t$. The feature visualization method is similar to prompt-to-prompt \cite{hertz2022prompt}.

From Fig.~\ref{fig:one_step}, we observe that constant weight exhibits artifacts and color shift in the first generation step, resulting in the final generated images with color and structure distortion. P2 and Min-SNR also show global artifacts and inconsistency in the first generation step. Thus, their generated images suffer from poor structures. On the contrary, our method is free of artifacts and color distortion in the whole generation process. 

From Fig.~\ref{fig:features}, we observe that constant, P2, and Min-SNR weight strategies struggle to generate clear facial architecture in the early generation steps. Besides, their intermediate feature maps of the final step also demonstrate poor global consistency. In contrast, our method demonstrates clear facial architecture even at very early steps, and the final feature maps are also more visually pleasing.

\section{Different Training Targets}   \label{ap:training targets}
In this section, we delve into the difference between $x_0$ prediction, $\epsilon$ prediction and $v$ prediction.
Most previous works \cite{nichol2021improved, dhariwal2021diffusion, nichol2021glide, rombach2022high} follow DDPM \cite{ho2020denoising} to predict the noise $\epsilon$. Some works \cite{salimans2021progressive, gu2022vector} use reparameterization to predict $x_0$. And some other works \cite{salimans2021progressive} employ the network to predict $v\equiv \alpha_t \epsilon - \sigma_t x_0$. $V$-prediction combines the strength of $x_0$-prediction and $\epsilon$-prediction, and is verified to be effective in sampling with fewer steps. 
Note that EDM \cite{karras2022elucidating} also realizes that directly predicting the Gaussian noise induces error amplification. While EDM resorts to precondition technique requiring network inputs and training targets to have unit variance. This operation is indeed in the same spirit as previous reparameterization method like $v$-prediction. Differently, our method comprehensively analyzes the bias issue in traditional $\epsilon$-prediction and propose the unbiased principle for solving error amplifying. Concretely, we formulate the bias issue with amplified error and systematically dive into the issue from its existence, impact, and reason, with quantitative and qualitative studies. 

Predicting different targets is mathematically equivalent. However, different prediction targets inherently correspond to different optimizing difficulty. $\epsilon$ prediction is theoretically easiest as the distribution of the optimizing target is simple and fixed. This also explains why predicting Gaussian noise $\epsilon$ with constant weight is most widely employed and becomes the de facto component of diffusion models. Significantly, this further validates the importance and meaning of our work unlocking the biased problem in $\epsilon$ prediction mode.

We also show the performance comparison of different training targets in CelebA-HQ dataset \cite{karras2017progressive} in Tab. \ref{table:targets}. Obviously, the performance of $\epsilon$-prediction is similar to that of $x_0$-prediction and $v$-prediction. 
While, our method can further substantially elevates the performance of $\epsilon$-prediction mode, and achieves the highest performance across different training targets.

\begin{table*}[ht]
\small
\setlength{\tabcolsep}{8pt}
\caption{Quantitative comparison of different training targets on CelebA-HQ dataset with $T=100$ steps.}
\label{table:targets}
\centering
\begin{tabular}{lcccccc}
	\toprule
        & $x_0$ & $v$ & $\epsilon$ & $\epsilon$ (P2) & $\epsilon$ (MinSNR) & $\epsilon$ (Ours)\\
        \midrule
	FID  & 12.3146 & 12.2773 & 12.0064  & 9.2972 & 9.3851 & 8.8363
    \\
	\bottomrule
\end{tabular}
\end{table*}

\section{More Visual Results}
\begin{figure*}[ht]
	\begin{center}
		\includegraphics[width=0.98\linewidth]{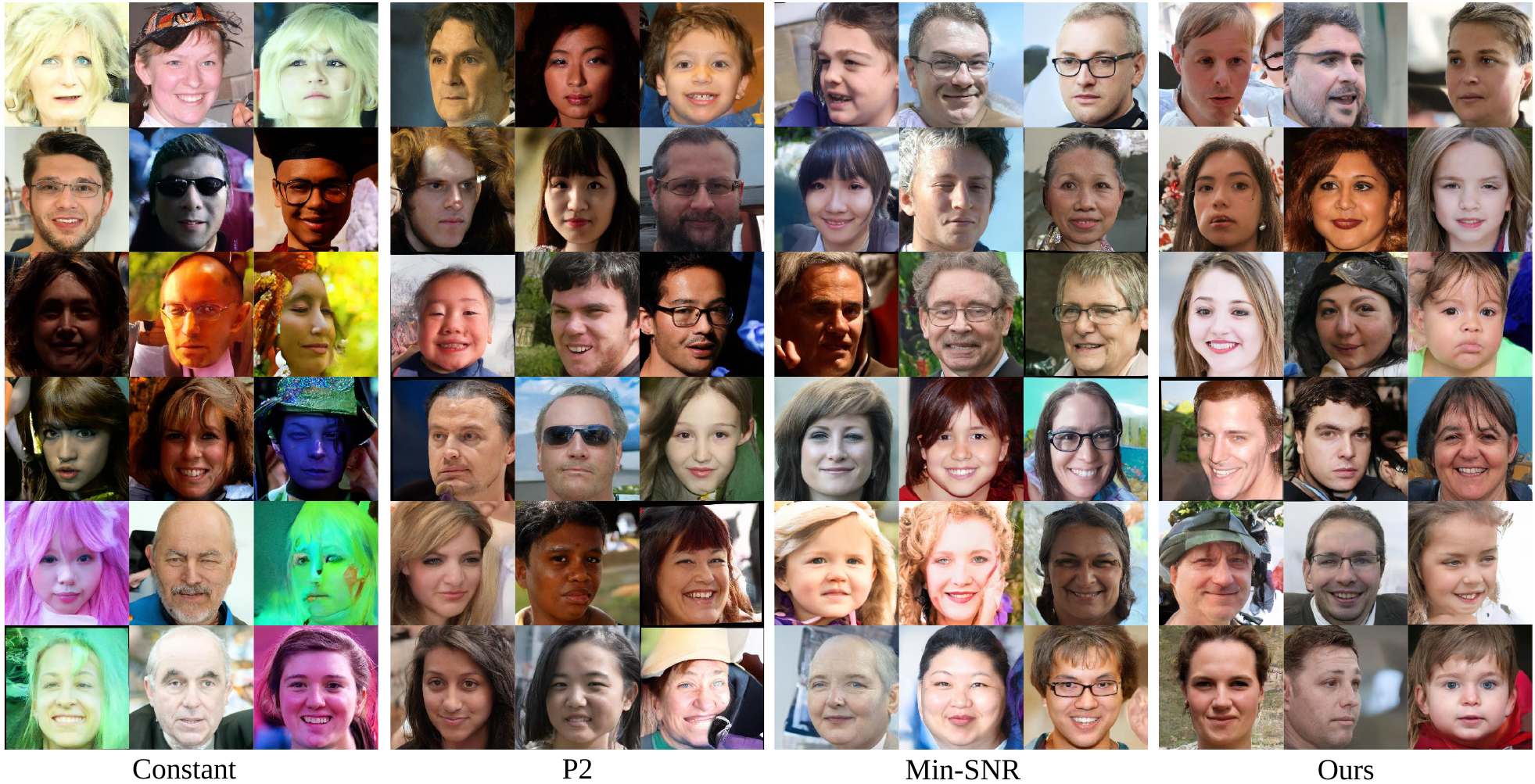}
	\end{center}
	\caption{More visual results on FFHQ dataset.}
	\label{fig:FFHQ}
\end{figure*}

\begin{figure*}[ht]
	\begin{center}
		\includegraphics[width=0.98\linewidth]{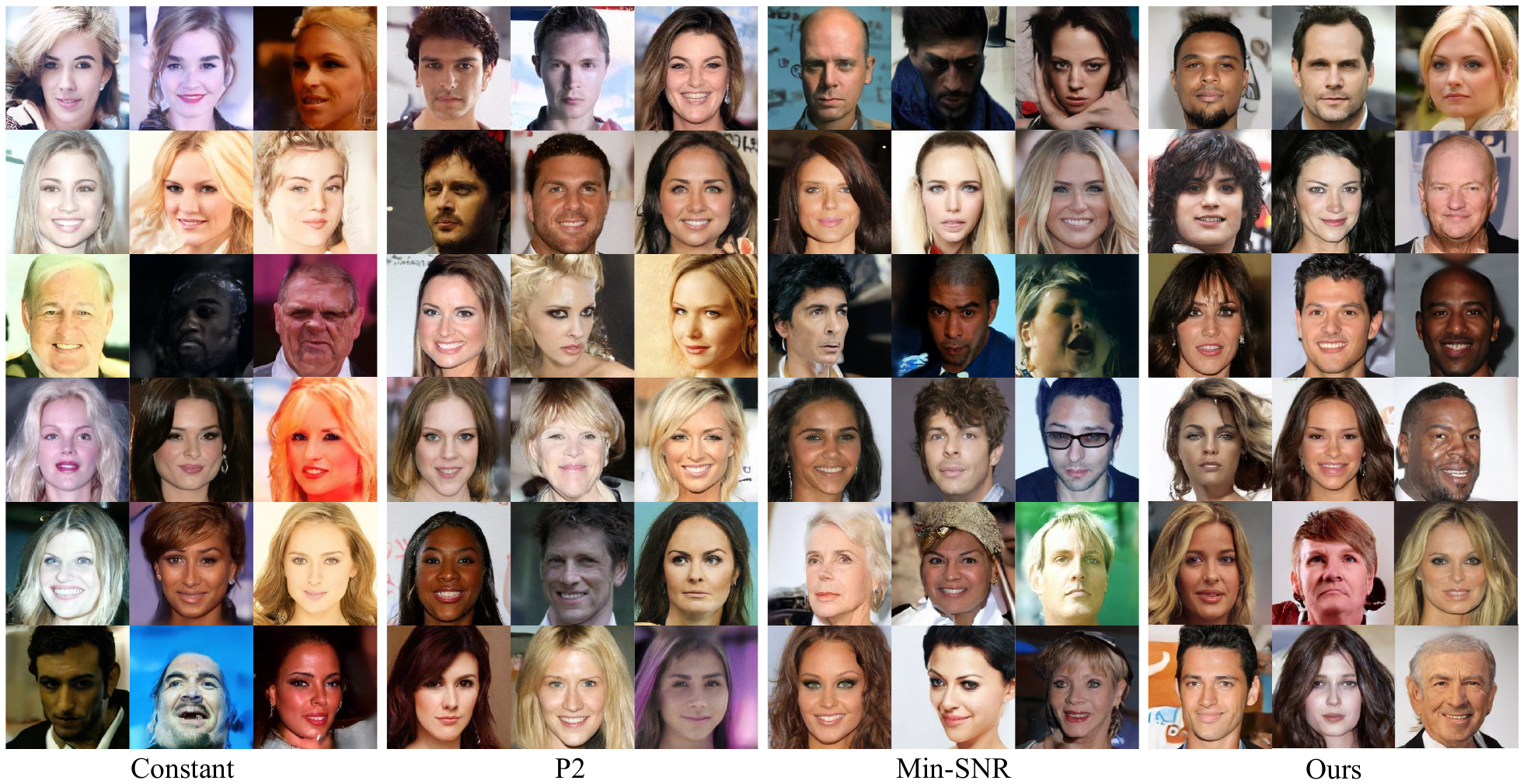}
	\end{center}
	\caption{More visual results on CelebA-HQ dataset.}
	\label{fig:CelebAHQ}
\end{figure*}

\begin{figure*}[ht]
	\begin{center}
		\includegraphics[width=0.98\linewidth]{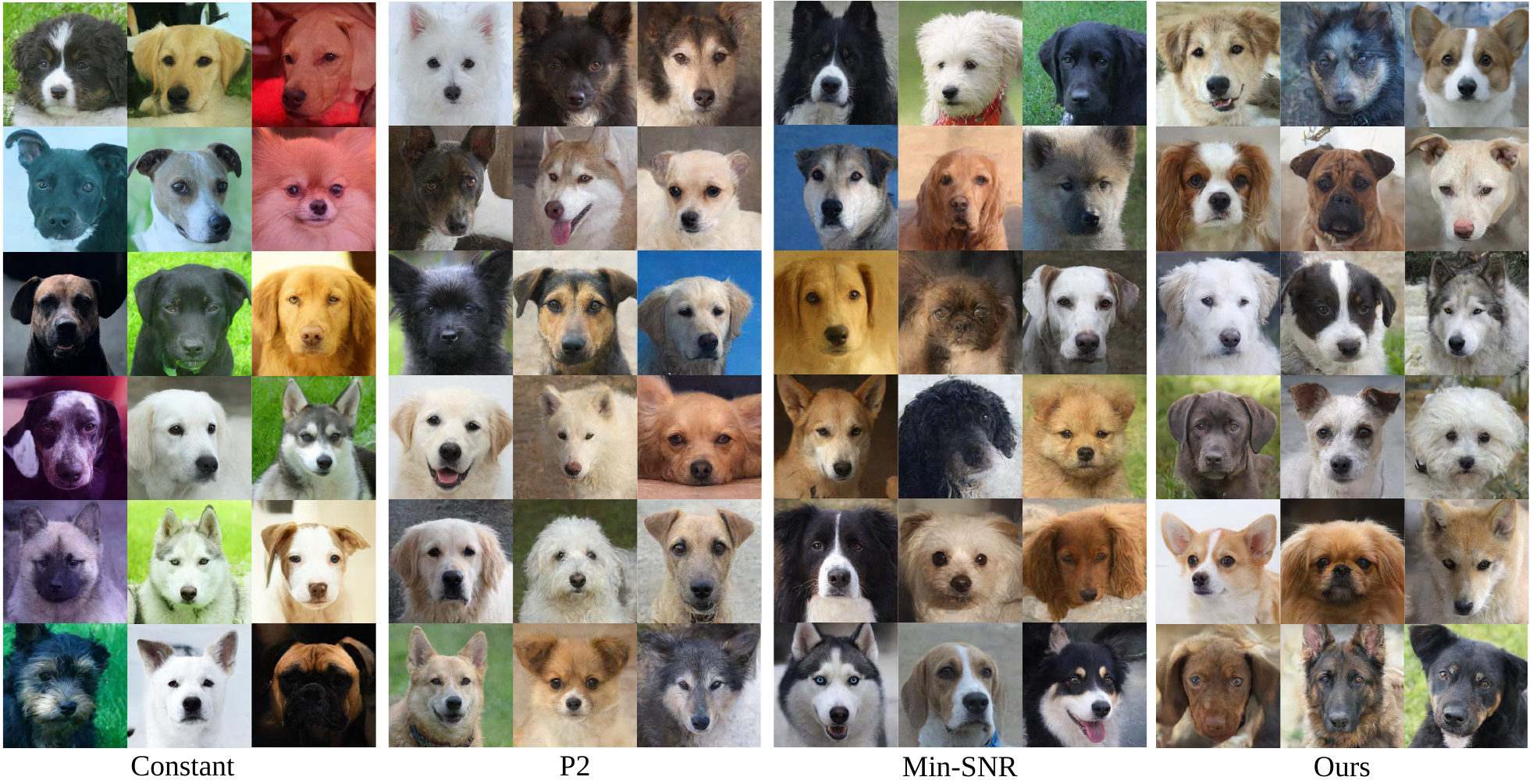}
	\end{center}
	\caption{More visual results on AFHQ-dog dataset.}
	\label{fig:AFHQ}
\end{figure*}

\begin{figure*}[ht]
	\begin{center}
		\includegraphics[width=0.98\linewidth]{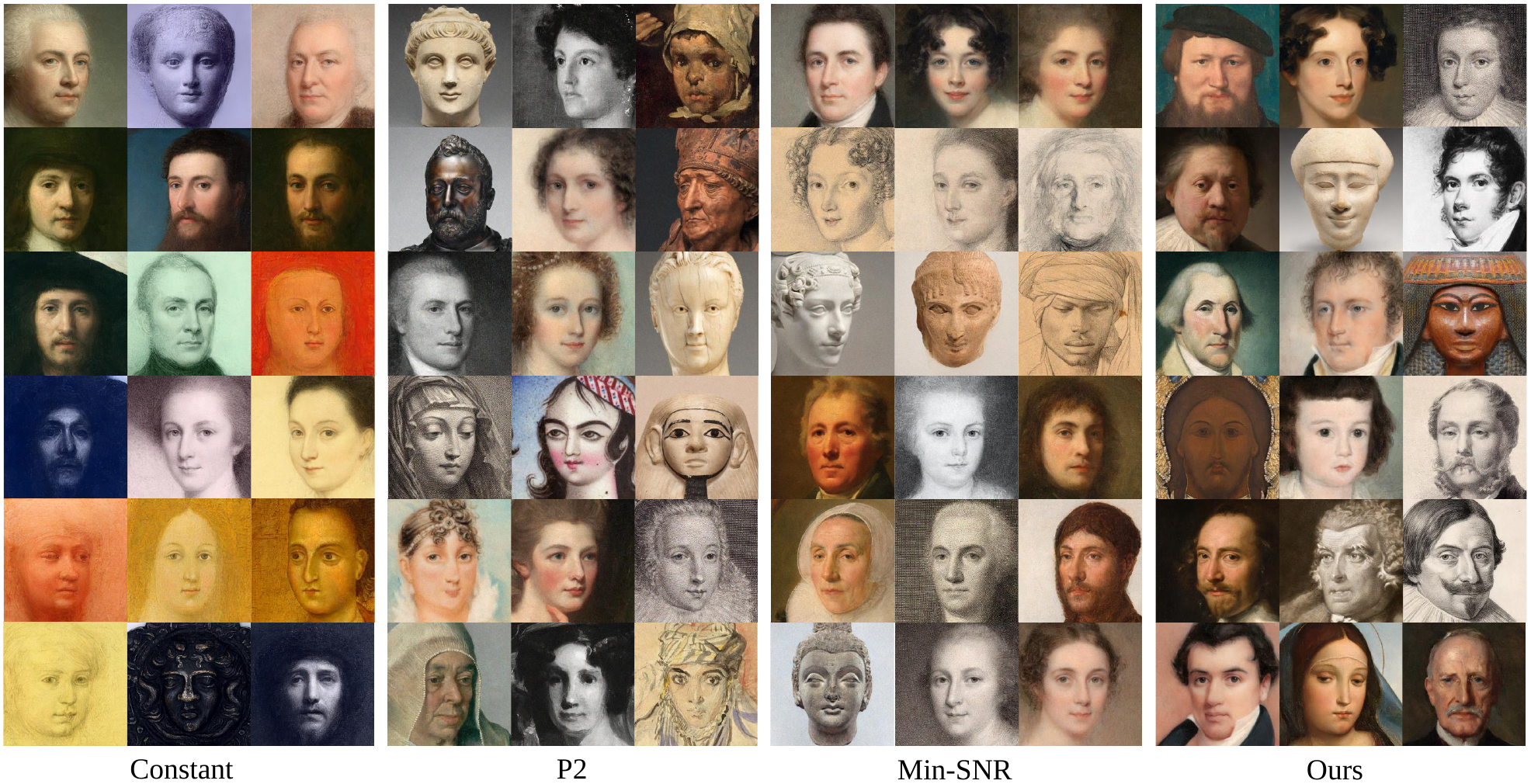}
	\end{center}
	\caption{More visual results on MetFaces dataset.}
	\label{fig:MetFaces}
\end{figure*}

In this part, we show more visual results of different weighting strategies on various datasets to further validate the effectiveness and robustness of our method. Fig.~\ref{fig:FFHQ}, Fig.~\ref{fig:CelebAHQ}, \ref{fig:AFHQ} , and \ref{fig:MetFaces} show the visual results on FFHQ \cite{karras2019style}, CelebA-HQ \cite{karras2017progressive}, AFHQ-dog \cite{choi2020stargan}, and MetFaces \cite{karras2020training} datasets, respectively.

\end{document}